\documentclass[twocolumn]{article}
\usepackage[utf8]{inputenc}
\usepackage{graphicx}
\usepackage[margin=1in]{geometry}
\usepackage{xcolor}
\usepackage{url}

\usepackage{sectsty}
\sectionfont{\fontsize{12}{15}\selectfont}

\title{Inferring and Improving Street Maps with Data-Driven Automation}
\author{Favyen Bastani\textsuperscript{1}, Songtao He\textsuperscript{1}, Satvat Jagwani\textsuperscript{1}, Edward Park\textsuperscript{1}, \\ Sofiane Abbar\textsuperscript{2}, Mohammad Alizadeh\textsuperscript{1}, Hari Balakrishnan\textsuperscript{1}, \\ Sanjay Chawla\textsuperscript{2}, Sam Madden\textsuperscript{1}, Mohammad Amin Sadeghi, \\
\textsuperscript{1}MIT CSAIL, \{favyen,songtao,satvat,parke,alizadeh,hari,madden\}@csail.mit.edu, \\
\textsuperscript{2}Qatar Computer Research Institute, \{sabbar, schawla\}@hbku.edu.qa}
\date{September 21, 2019}

\begin{document}

\maketitle

\section{Introduction}

Street maps help to inform a wide range of decisions. Drivers, cyclists, and pedestrians rely on street maps for search and navigation. Rescue workers responding to disasters like hurricanes, tsunamis, and earthquakes rely on street maps to understand where people are, and to locate individual buildings~\cite{puertorico}. Transportation researchers rely on street maps to conduct transportation studies, such as analyzing pedestrian accessibility to public transport~\cite{pedaccess}.
%And last but not least, with the impending arrival of autonomous vehicles,
Indeed, with the need for accurate street maps growing in importance, companies are spending hundreds of millions of dollars to map roads globally\footnote{See e.g. ``Uber Will Spend \$500 Million on Mapping to Diverge From Google'', Fortune (2016-07-15) by Kirsten Korosec.}.
%It has been reported that the Google Maps team employs over 7,000 people to maintain the application~\cite{google7000}.

%\sanjay{Do we really need to motivate  why we need street maps ? Instead we should motivate why we need  cost-efficient procedures to update maps}

However, street maps are incomplete or lag behind new construction in many parts of the world. In rural Indonesia, for example, entire groups of villages are missing from OpenStreetMap, a popular open map dataset~\cite{roadtracer}. In many of these villages, the closest mapped road is miles away. In Qatar, construction of new infrastructure has boomed in preparation for the FIFA World Cup 2022. Due to the rapid pace of construction, it often takes a year for digital maps to be updated to reflect new roads and buildings.\footnote{An example of a subdivision in Doha, Qatar that was missing from maps for years is detailed at  \url{https://productforums.google.com/forum/\#!topic/maps/dwtCso9owlU}.} Even in countries like the United States where significant investments have been made in digital maps, construction and road closures often take days or weeks to be incorporated in map datasets.

% todo: business spending a lot of money, applications like autonomous vehicles make it mission critical, the data is proprietary -> have openstreetmap
% we make it cheaper for businesses and also enable more open data
% * add # map editors from intro version 2 (done)
% * add some stuff about doha like in the footnotes in our papers (done)
% maybe put saikat as reviewer (done)

% todo: put evaluation results in the sections
% highlight that we have results on many cities and that we trained on different cities

% for roadrunner/roadtracer, can mention precision/recall in the paper but show qualitative results

These problems arise because the processes used to create and maintain street maps today are extremely labor-intensive. Modern street map editing tools allow users to trace and annotate roads and buildings on a map canvas overlayed on relevant data sources, so that users can effectively edit the map while consulting the data. These data sources include satellite imagery, aerial imagery, and GPS trajectories (which consist of sequences of GPS positions captured from moving vehicles).
%\favyen{I think aerial imagery is always from aircraft, and satellite imagery is always from satellite.}
Although the data presented by these tools help users to update a  map dataset, the manual tracing and annotation process is cumbersome and a major bottleneck in map maintenance.
%\sanjay{We should distinguish between primary data sources and secondary data sources. Google collects primary data for map construction. Can we use secondary data to update maps.}
%\favyen{What is an example of secondary data source? I think we only use primary source (GPS, satellite imagery).}
%\sanjay{Primary and secondary has a different definition. Primary data means that the data was collected for a specific task. For example Google cars are sent out to collect data to build Google Maps. But we use secondary GPS data which was not necessarily collected for map inference but we use it for that purpose. The basic point is that secondary data is cheaper to acquire and if we can use it to build maps, the cost of map making goes down.}

Over the past decade, many \emph{automatic map inference} systems have been proposed to automatically extract information from these data sources at scale. Several approaches develop unsupervised clustering and density thresholding algorithms to construct road networks from GPS trajectory datasets~\cite{ahmed2015comparison,biagioni2012inferring,cao2009gps,liu2012mining,kharita}. Others apply machine learning methods to process satellite imagery and extract road networks~\cite{mnih2010learning,wegner2015road}, building polygons~\cite{hamaguchi2018cvpr,zhao2018cvpr}, and road attribute annotations (e.g., the number of lanes, presence of cycling lanes, or presence of street parking on a road)~\cite{cadamuro2018assigning,mattyus2015enhancing}.

However, automatic map inference has failed to gain traction in practice due to two key limitations. First, existing inference methods have high error rates (low precision), which manifest in noisy outputs containing incorrect roads, buildings, and attribute annotations. Second, although prior work has shown how to detect roads and buildings from the data sources, the challenge of leveraging this information to update real street map datasets has not been addressed. We argue that even with lower error rates, the quality of outputs from automatic approaches is below that of manually curated street map datasets, and semi-automation is needed to efficiently but robustly take advantage of automatic map inference to accelerate the map maintenance process.
%\sanjay{The second key limitation is not very clear. Could you elaborate or rewrite. }
%\favyen{Thanks, I expanded on it a bit, is it better?}

\begin{figure}[t!]
\begin{center}
	\includegraphics[width=\linewidth]{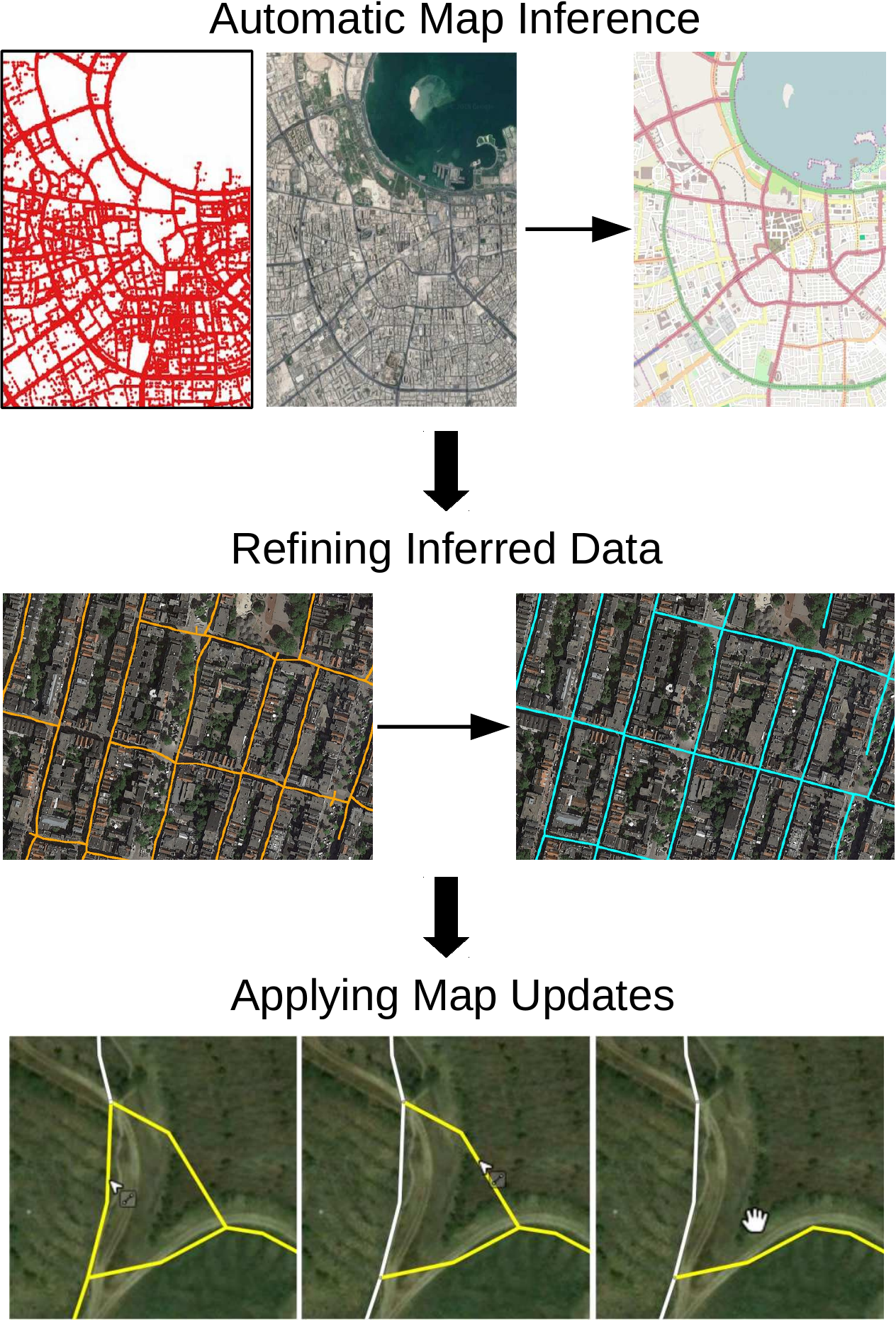}
\end{center}
	\caption{Overview of the Mapster map editing system. We first infer road network from satellite imagery and GPS data. Then, we transform the map to make it look more realistic. Finally, we have an interactive system to apply map updates.}
\label{fig:outline}
\end{figure}

% todo: unifying theme: eliminate post-processing where possible
% by directly producing graph / attributes
% only produce outputs when there is reasonable confidence and also GAN help to make it look good or idk

At MIT and QCRI, we have developed a number of algorithms and approaches to address these challenges~\cite{maid, roadtracer,roadrunner},
%\srm{cite all the papers we wrote}
which we combined into a new system we call {\it Mapster}. Mapster is a human-in-the-loop street map editing system that incorporates three components to robustly accelerate the mapping process over traditional tools and workflows: high-precision automatic map inference, data refinement, and machine-assisted map editing.

First, Mapster applies automatic map inference algorithms to extract initial estimates from the raw data sources. Although these estimates are noisy, we minimize errors by applying two novel approaches, iterative tracing for road network inference and graph network annotation for attribute inference, that replace heuristic, error-prone post-processing steps present in prior work with end-to-end machine learning and other more robust methods.

Second, Mapster refines the noisy estimates from map inference into map update proposals by removing several types of errors and reducing noise. To do so, we apply conditional generative adversarial networks (cGANs) trained to transform the noisy estimates into refined outputs that are more consistent with human-annotated data.

Finally, a machine-assisted map editing framework enables the rapid, semi-automated incorporation of these proposed updates into the street map dataset. This editing tool addresses the problem of leveraging inferred roads, buildings, and attribute annotations to update existing street map datasets.

Figure \ref{fig:outline} summarizes the interactions between these components.
%\sanjay{We should briefly explain our design choice for having a three-step approach and why we are settling for a semi-automatic human-in-the-loop framework and why is it not fully-automatic.  Since "iterative tracing" is the key novelty, we should put it in different places. Maybe also in the Figure caption.} \favyen{alright, edited the text a bit.}
%Mapster addresses high error rates through the introduction of a separate data refinement stage and improvements to automatic map inference algorithms. We propose iterative tracing to increase the accuracy of road network inference by accounting for context in the input data sources and topology in the output roads when inferring a road network. Similarly, to reduce errors in road attribute annotations, 
%we develop a hybrid neural network architecture that combines a convolutional neural network with a graph neural network in a jointly trained model.
%we develop \songtao{How about: we remove the error-prone post-processing step present in prior work by developing} a hybrid neural network architecture that combines a convolutional neural network with a graph neural network in a jointly trained model.
We include links to two videos demonstrating the execution of Mapster, along with a link Mapster's source code (which we have released as free software), in the footnote.\footnote{Video of iterative tracing in action: \url{https://youtu.be/3_AE2Qn-Rdg}. Video of machine-assisted map editing: \url{https://youtu.be/i-6nbuuX6NY}. Mapster source code: \url{https://github.com/mitroadmaps}.}

Below, we first introduce our automatic map inference approaches for inferring roads and road attributes, which obtain substantially higher precision than prior work. We then detail our data refinement strategy, which applies adversarial learning to improve the quality of inferred road networks. Finally, we discuss our machine-assisted map editor, which incorporates novel techniques to maximize user productivity in updating street maps. We conclude with a discussion of future work.

\section{Automatically Inferring Road Networks}

Given a base road network, which may be empty or may correspond to the roads in the current street map dataset, the goal of road network inference is to leverage GPS trajectory data and satellite imagery to produce a road network map that covers roads not contained in the base map. The road network map is represented as a graph where vertices are annotated with spatial longitude-latitude coordinates, and edges correspond to straight-line road segments.

\begin{figure*}[t!]
\begin{center}
	\includegraphics[width=\linewidth]{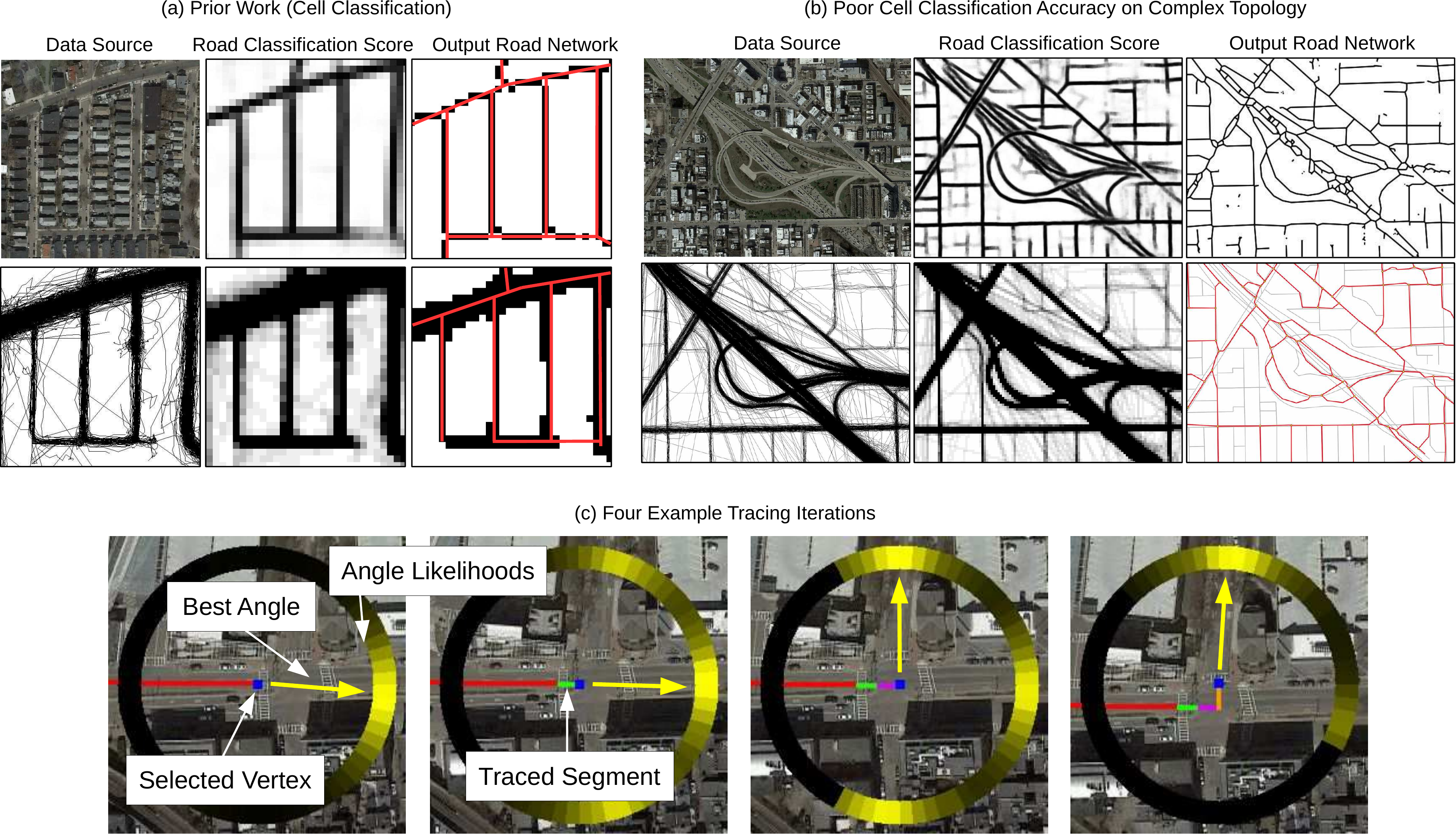}
\end{center}
	\caption{(a) Prior automatic mapping approaches operating on satellite imagery (top) and GPS trajectories (bottom). In the center, we show the per-cell classification scores, and on the right, the result of thresholding and cell connection. (b) These approaches produce noisy outputs around complex road topology like highway interchanges. (c) Iterative tracing on satellite imagery. A CNN predicts the likelihood that there is a road at each of 64 angles from a vertex; higher likelihoods from the blue vertex are shown in yellow in the outer circle, and lower likelihoods in black. Iterative tracing gradually expands the coverage of the map: on each iteration, it adds a segment corresponding to the highest likelihood vertex and direction.}
\label{fig:tracing}
\end{figure*}

Broadly, prior approaches infer roads by dividing the space into a 2D grid, classifying whether each grid cell contains a road, and then connecting cells together to form edges. Figure \ref{fig:tracing}(a) summarizes this strategy. For satellite imagery, recent work obtains the per-cell classification by applying deep convolutional neural networks (CNNs) that segment the imagery, transforming the input imagery into a single-channel image indicating the neural network's confidence that there is a road at each pixel~\cite{cheng2017automatic,costea2016aerial,deeproadmapper}. For GPS trajectories, several approaches perform the classification based on the number of GPS trajectories passing through each cell~\cite{biagioni, chen2008, davies, wenhuan2009}.

However, we find that these methods exhibit low accuracy when faced with practical challenges such as noisy data and complex road topology. Figure \ref{fig:tracing}(b) shows the outputs of prior work around a major highway junction in Chicago. Noise in the per-cell classification estimates are amplified when we connect cells together to draw edges, resulting in garbled road network maps.

Indeed, both GPS trajectory data and satellite imagery exhibit several types of noise that make robust identification of roads challenging. While GPS samples in open areas are typically accurate to four meters, in practice, due to high-rise buildings and reflection, GPS readings may be as far off as tens of meters. Correcting this error is difficult because errors are often spatially correlated --- multiple GPS readings at the same location may be offset in the same way as they encounter the same reflection and distortion issues. Similarly, roads in satellite imagery are frequently occluded by trees, buildings, and shadows. Furthermore, distinguishing roads and buildings from non-road trails and surface structures in imagery is often nontrivial.

%\songtao{Should we say something about why we don't choose to keep improving the segmentation or holistic approach (e.g., the fundamental limitation of them?), instead we choose to pursue a new direction - iterative tracing? Or maybe the answer has been already in the introduction. }
%\favyen{I updated the second sentence to focus more on this}
To substantially improve precision, we adopt an iterative road tracing approach in lieu of the per-cell classification strategy.
Our iterative tracing method mimics the gradual tracing process that human map editors use to create road network maps, thereby eliminating the need for the heuristic post-processing steps that prior work applies to draw edges based on cell classification outputs.

Iterative tracing begins with the base map, and on each iteration, it adds a single road segment (one edge) to the map. To decide where to position this segment, it uses the data source to compute two values for each vertex in the portion of the map traced so far: (a) a confidence that an unmapped road intersects the vertex, and (b) the most likely angular direction of the unmapped road. It then selects the vertex with the highest confidence, and adds a segment in the corresponding direction. This procedure is repeated, adding one segment at a time to the map, until the highest confidence for the presence of an unmapped road falls below a threshold. We illustrate the iterative tracing procedure in Figure \ref{fig:tracing}(c).
%\sanjay{The figure is not very clear.  The yellow circumference is getting bigger but then shrinks in the fourth subfigure.}
%\favyen{Improved the figure.}

We develop different approaches to compute the unmapped road confidence and direction from satellite imagery~\cite{roadtracer} and from GPS trajectories~\cite{roadrunner}. With satellite imagery, we compute these through a deep neural network. We develop a CNN model that inputs a window of satellite imagery around a vertex of the road network, along with an additional channel containing a representation of the road network that has been traced so far. We train the CNN to output the likelihood that an unmapped road intersects the vertex, and the most likely angular direction of this unmapped road.

%\begin{figure}[t]
%\begin{center}
%	\includegraphics[width=\linewidth]{figures/roadrunner}
%\end{center}
%	\caption{When performing iterative tracing on GPS trajectories, we consider the geometry of the current road traced so far when extending that road during the tracing process.}
%\label{fig:roadrunner}
%\end{figure}

% todo: some bold text here for GPS / satellite

With GPS trajectories, we compute the values at a vertex based on the peak direction of trajectories that pass through the vertex. We identify all trajectories that pass through the vertex, and construct a smoothed polar histogram over the directions that those trajectories follow after moving away from the vertex. We then apply a peak finding algorithm to identify peaks in the histogram that have not already been explored earlier in the tracing process. We select the peak direction, and measure confidence in terms of the volume of trajectories that match the peak direction.

%Importantly, when identifying trajectories that pass through a vertex, we consider not only the position of that vertex, but also match trajectories with predecessor vertices. This effectively excludes trajectories that correspond to nearby but separate roads, such as overpasses and underpasses, thereby ensuring that these roads remain correctly disentangled in the output road network map.

%\sanjay{We should mention that we can combine the inference from satellite and GPS data and that often leads to a more complete map.}
%\favyen{Added a paragraph below.}

Often, both satellite imagery and GPS trajectory data may be available in a region requiring road network inference. We develop a two-stage approach that leverages both data sources when inferring road networks to reduce errors and improve the map quality. In the first stage, we run iterative tracing using GPS data to infer segments along high-throughput roadways. Because these roads have high traffic volume, they are covered by large numbers of GPS trajectories, and so can be accurately traced with GPS data. At the same time, junctions along high-throughput roads (especially controlled-access highways) are generally more complex, often involving roundabouts and overpasses. These features make tracing based on satellite imagery challenging.

In the second stage, we fill in gaps in the road network with lower-throughput residential and service roads that were missed in the first stage by tracing with satellite imagery. These roads have simple topology and are covered by few GPS trajectories, making imagery a preferred data source.

%We perform two evaluations to measure the effectiveness of the Mapster system for accurately inferring street map data and accelerating street map maintenance. These evaluations cover both iterative tracing and inferred data refinement.

%We conduct the evaluation in four urban areas where we have both satellite imagery and GPS trajectory data: Boston, Chicago, Los Angeles, and New York. We measure accuracy in terms of a precision-recall score, where we use OpenStreetMap data as a ground truth dataset of correct roads. Precision is the fraction of inferred roads that are correct, and recall is the fraction of correct roads that were inferred.
%\sanjay{Can we include some Doha results. Maybe qualitative.}

\begin{figure*}[t!]
\begin{center}
	\includegraphics[width=0.75\linewidth]{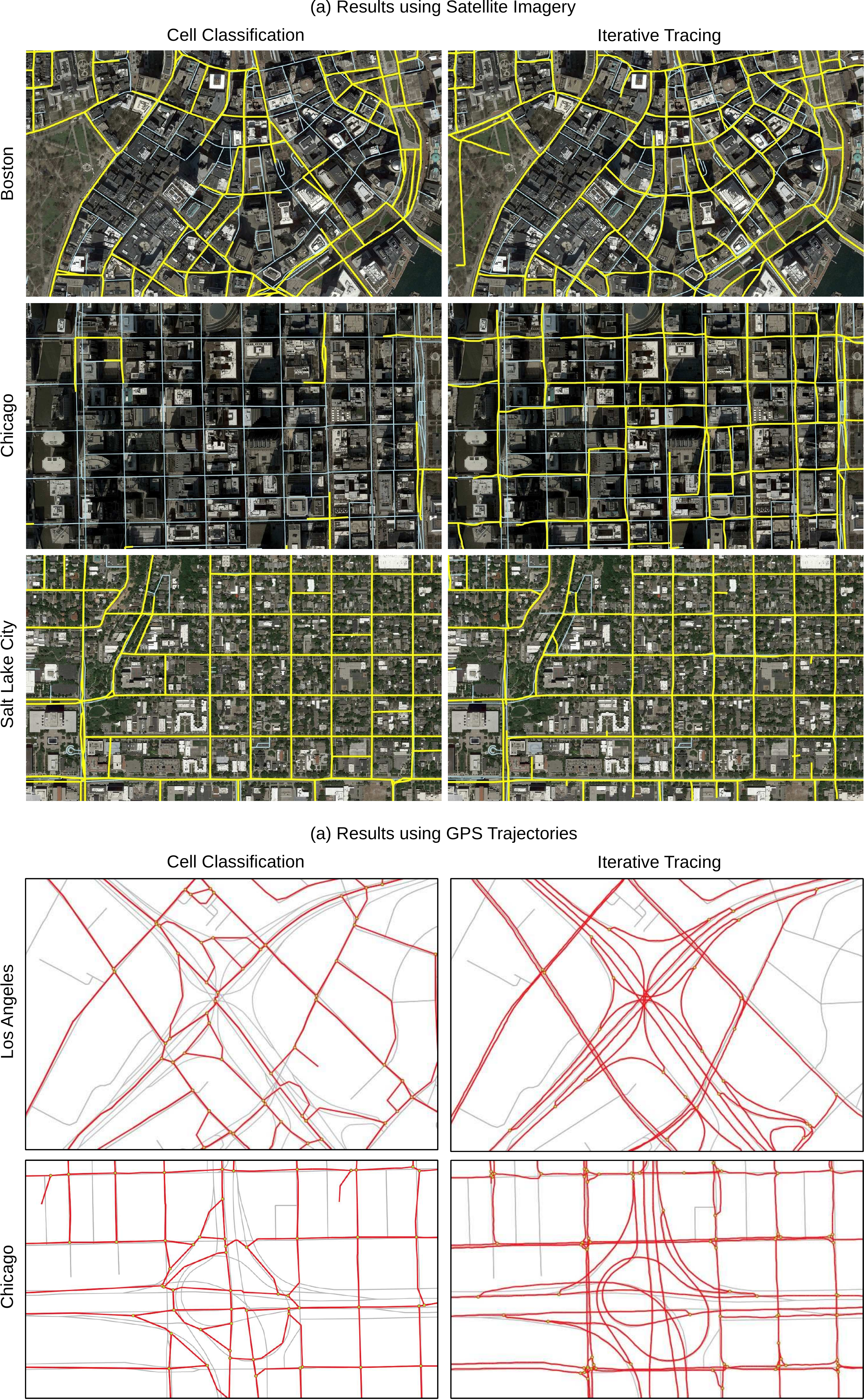}
\end{center}
	\caption{Qualitative results comparing iterative tracing to prior cell classification approaches. (a) shows road networks inferred from satellite imagery, and (b) shows road networks inferred from GPS trajectories. We show inferred roads in the foreground and OpenStreetMap data in the background.}
\label{fig:results}
\end{figure*}

We evaluate our method by comparing road networks inferred through iterative tracing against those inferred by prior cell-classification approaches. Figure \ref{fig:results} shows qualitative results using satellite imagery in Boston, Chicago, and Salt Lake City, and using GPS trajectory data in Chicago and Los Angeles. In contrast to cell classification, iterative tracing from satellite imagery infers roads robustly despite occlusion by buildings and shadows in dense urban areas. In lower density areas like Salt Lake City, iterative tracing performs comparably to prior work.

Cell classification from GPS trajectories produces noisy outputs at crucial but complex map features like highway interchanges. Despite the intricate connections at these features, iterative tracing accurately maps the interchanges.

We show quantitative results in \cite{roadtracer, roadrunner}.

%Figure \ref{fig:tracing-results}(a) show the precision-recall curves for cell-classification and iterative tracing on satellite imagery and GPS trajectories, averaged over the four cities. Our iterative tracing approach yields higher precision for the same recall in almost all cases. Although iterative tracing with satellite imagery is unable to produce maps with better than 46\% recall, cell-classification is only able to obtain higher recalls with significant error rates (low precision). Inferring high-precision road networks is crucial for ensuring that inferred roads can be efficiently integrated with the existing street map dataset: each inferred road containing errors slows down the machine-assisted map editing process because fixing these errors is much more time-consuming than validating a correct road.
%\sanjay{Very important point}

%In Figure \ref{fig:results}(b), we highlight results in Doha, Qatar from our two-stage road network inference technique, where we leverage both GPS data and satellite imagery to trace roads with high accuracy. We accurately trace major roads with complex topology using GPS trajectory data, and trace minor roads with more straightforward topology using satellite imagery.

\section{Inferring Road Attributes}
%\songtao{changing this part to focus on the potential theme - 'removing error-prone post-processing'}
Modern navigation systems use more than just the road topology -- they also make use of a number of road metadata attributes,  such as the number of lanes, presence of cycling lanes, or the presence of street parking along a road.  As a result, inferring these attributes is an important part of the Mapster system.

%Prior work in road attribute inference applies an image classification approach that trains a convolutional neural network (CNN) to predict the road attributes given a window of satellite imagery around some location along the road.

Prior work in road attribute inference applies an image classification approach that trains a convolutional neural network (CNN) to predict the road attributes given a window of satellite imagery around some location along the road. Then the local prediction at each location is post-processed through a global inference phase, e.g., inference in a Markov Random Field (MRF), to remove scattered errors. 

\begin{figure}[t!]
\begin{center}
	\includegraphics[width=\linewidth]{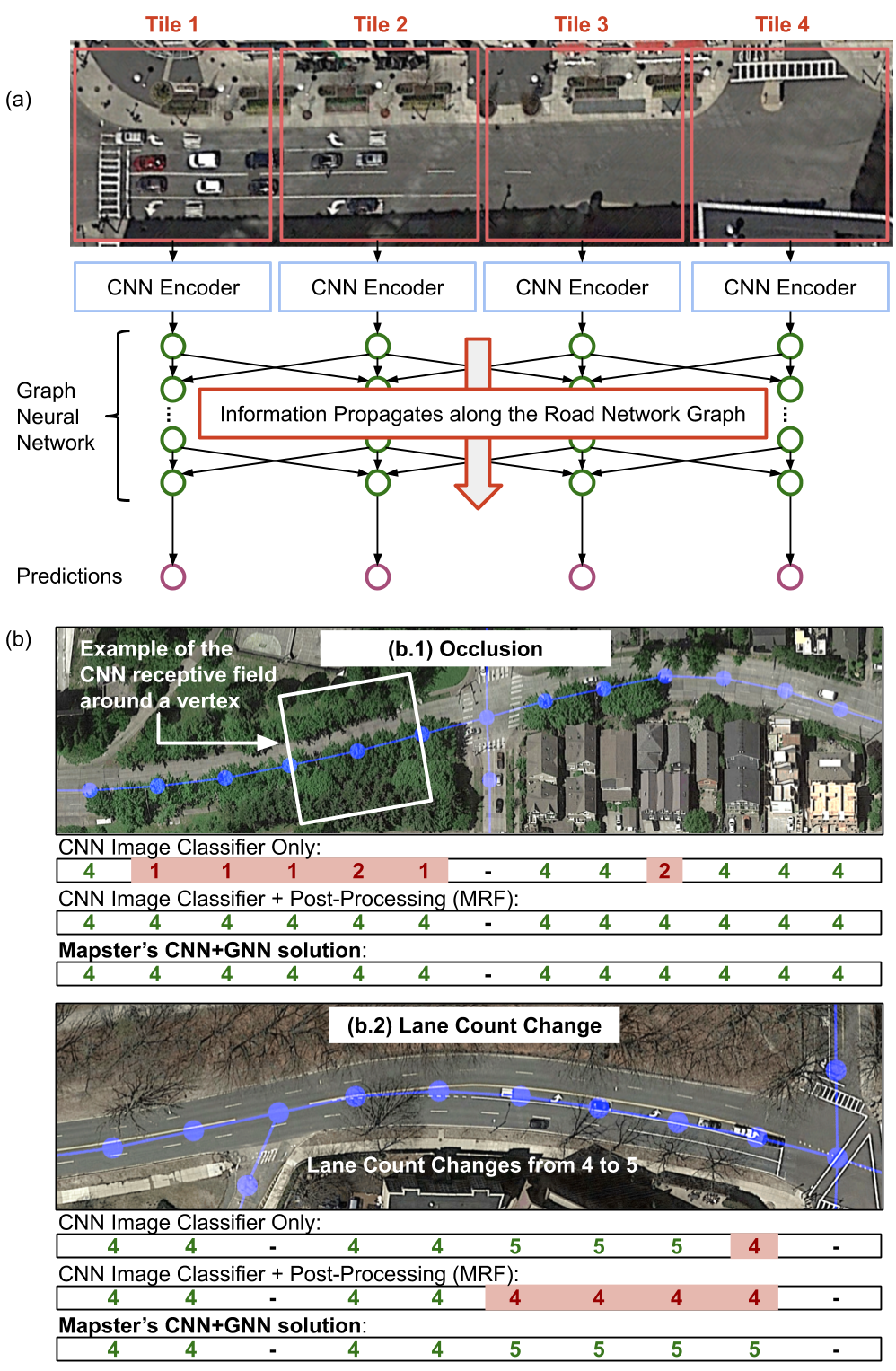}
\end{center}
	\caption{(a) The hybrid neural network architecture proposed in this work. (b) Examples on inferring the number of lanes. In each image, blue lines show the road graph. The number of lanes predicted by the CNN Image Classifier and Mapster on each segment are shown along the bottom of each figure. We color the output numbers green for correct predictions and red for incorrect predictions.}
\label{fig:roadannotator}
\end{figure}

This global inference phase is necessary because the CNN prediction at each location is often erroneous due to the \textit{limited receptive field} of the CNN -- in many cases, local information in the input window of satellite imagery is not sufficient to make a correct prediction. For example, in Figure \ref{fig:roadannotator}(b.1), the road on the left side is occluded by trees. If the CNN inputs a window only from the left part of the road, it will be unable to correctly determine the number of lanes. The global inference phase algorithm can take all the predictions along the road as well as prior knowledge, such that the road attributes are often homogeneous along the road, into account to correct the errors in CNN classifiers.

However, we find this post-processing fix is often error-prone. For example, see Figure \ref{fig:roadannotator}(b.2), where the lane count changes from 4 to 5 near an intersection. The image classifier outputs partially incorrect labels. However, the post-processing strategy cannot fix this problem as the global inference phase only takes the predictions from the image classifier as input and it may not be able to tell whether the number of lanes indeed changes or it is an error of the image classifier. This limitation is caused by the \textit{information barrier} induced by the separation of local classification and global inference; the global inference phase can only use the image classifier's prediction as input, but not other important information such as whether trees occlude the road or whether the road width changes.  

To overcome the information barrier limitation, we develop a hybrid neural network architecture (see Figure \ref{fig:roadannotator}(a)) that combines a CNN with a graph neural network (GNN). The CNN extracts local features from each segment along a road. The GNN then propagates these features along the road network. 
The end-to-end training of the combined CNN and GNN model is the key to the success of the method: our algorithm doesn't rely on the often handcrafted and error-prone post-processing algorithm; instead, we learn the post-processing rules as well as the required CNN features
%for the post-processing rules 
directly from the data.
The usage of the GNN in our system eliminates the \textit{receptive field limitation} of local CNN image classifiers and the combination of CNN and GNN eliminates the \textit{information barrier limitation} present in prior work, allowing the model to learn complex inductive rules that make it more robust to challenges such as occlusion in satellite imagery and the partial disappearance of important information (Figure \ref{fig:roadannotator}(b)).

\section{Refining Inferred Data}

\begin{figure*}[t!]
\begin{center}
	\includegraphics[width=0.9\linewidth]{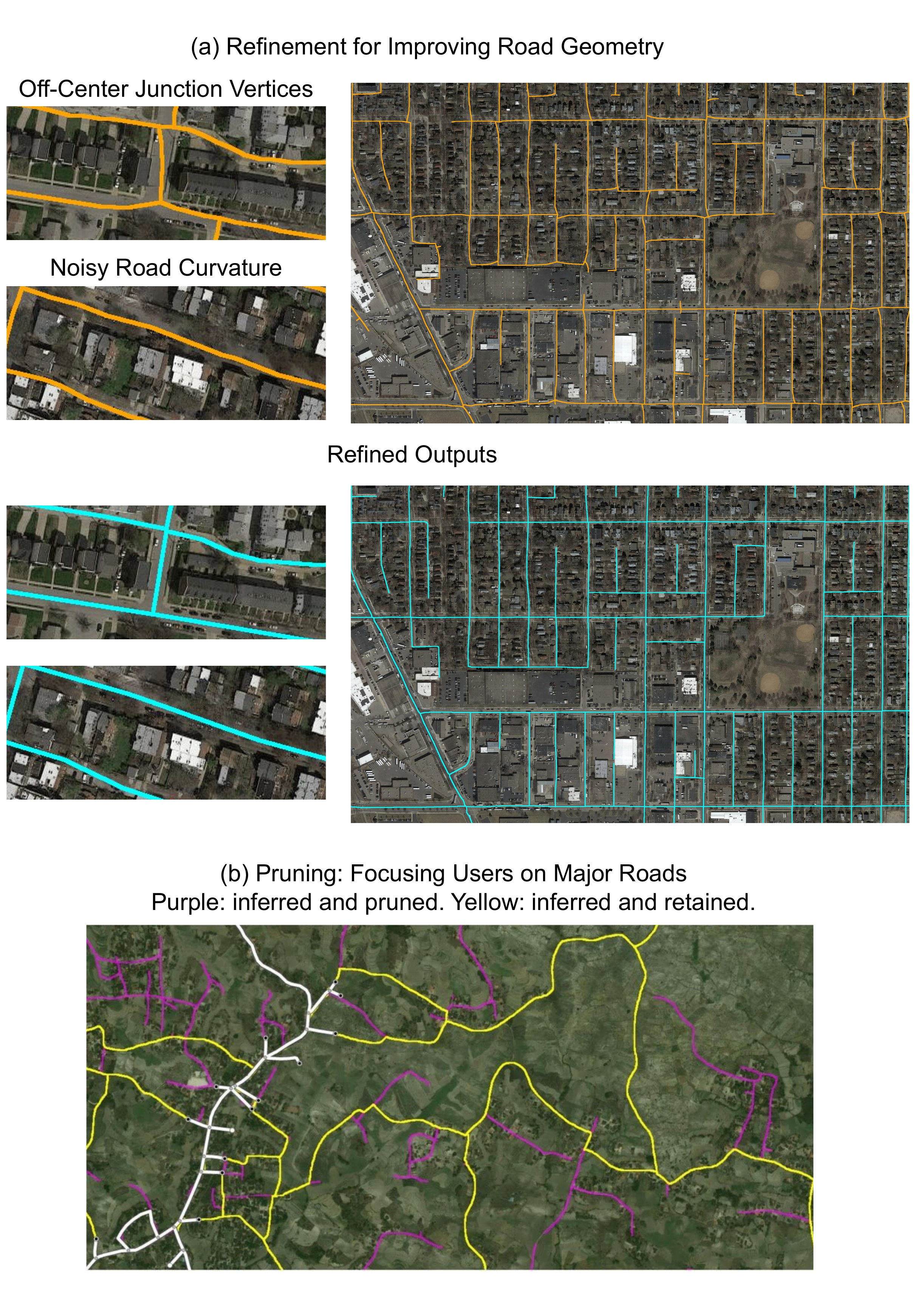}
\end{center}
	\caption{(a) Road networks before refinement, shown in orange, contain geometrical abnormalities. The refined road networks, shown in blue, clean up these noisy features.}
\label{fig:refinement}
\end{figure*}

Although iterative tracing improves substantially over prior grid cell classification approaches, it nevertheless creates road networks with noisy features that clearly distinguish them from human-drawn maps. In particular, iterative tracing effectively captures the topology of road networks, but leaves geometrical abnormalities such as the examples of off-center junction vertices and noisy road curvature in Figure \ref{fig:refinement}(a).

To refine the map and rectify these abnormalities, we use a model to learn the realistic road appearance. Our model is based on conditional generative adversarial networks (cGANs)~\cite{gan}. These networks learn to realistically reproduce complex image-to-image transformations, and have been successfully applied for many tasks including adding color to black-and-white images and transforming photos taken in daylight to plausible nighttime photos of the same scene~\cite{pix2pix}.
%\sanjay{Why CGANs and not just GANs ? }
%\favyen{There was a version where we introduced GANs first, and then cGANs, but then there is too much background.}

An obvious transformation to learn for map inference is transforming satellite imagery or representations of GPS trajectories into road networks, where the output image contains lines corresponding to road segments. However, we find that the learning problem is too difficult under this strategy, and the cGAN model fails to learn to robustly identify roads --- instead, it learns primarily to produce arbitrary lines that resemble lines in the ground truth road network representation.

Thus, instead, our cGAN model inputs not only satellite imagery or GPS trajectory data, but also a representation of the road network produced by iterative tracing. It outputs a refined road network representation where abnormalities in the input network have been corrected. By providing this initial road network, we reduce the complexity of the transformation, and thereby assist the cGAN to learn the transformation, especially early in the training process. Incorporating the initial road network representation derived from iterative tracing into the cGAN was a crucial insight that made training the adversarial model feasible.
%\sanjay{The fact that we included the initial representation from iterative tracing into cGAN was an important insight.  An advantage of GANs is that we don't need parallel sets (corpus), i.e., as long as we have a set of "true maps" that should be sufficient for a discriminator.} %\favyen{incorporated the first sentence. for the way we trained it we still do need paired input-output examples.}

Our cGAN architecture consists of two components: a \emph{generator} that produces refined road networks given an initial road network and a data source, and a \emph{discriminator} that learns to distinguish refinements made by the generator from road networks in the ground truth dataset. This network is adversarial because we train the generator and discriminator with opposing loss functions: the discriminator minimizes its classification error at distinguishing ground truth and generated (refined) road networks. In contrast, the generator learns by maximizing the discriminator's classification error. Thus, in effect, we train the generator by having it learn to fool the discriminator into classifying its generated road network as ground truth.

The initial road network provided to the generator enables the cGAN model to learn to produce realistic road networks. At first, the generator simply copies the road network from its input to its output in order to deceive the discriminator. However, once the discriminator learns to better distinguish the iterative tracing road networks from hand-drawn ground truth roads, the generator begins making small adjustments to the roads so that they appear to be hand-drawn. As training continues, these adjustments become more robust and complete.

Figure \ref{fig:refinement}(a) shows the outputs of our cGAN in blue, both on the geometrical abnormalities introduced earlier and on a larger region in Minneapolis. While refinement does not substantially alter the topology of the road network, the cGAN improves the geometry so that inferred roads resemble hand-mapped roads. These geometry improvements help to reduce the work needed to integrate inferred data into the street map.
%\sanjay{Overall I really like this section!}

%Second, we highlight qualitative results from inferred data refinement in Figure \ref{fig:results}(c) to demonstrate how refinement improves the quality of inferred road networks. While refinement does not substantially alter the topology of the road network (and thus has little impact on the precision-recall metric), the generative adversarial network and other refinement steps improve the geometry so that the inferred roads resemble hand-mapped roads. As with increased precision in automatic map inference, these geometry improvements help to reduce the work needed to integrate inferred data into the street map. Rather than requiring users to shift junctions and straighten roads manually, the data refinement stage automatically performs these adjustments in most cases. As the qualitative results show, Mapster's cGAN-based refinement approach succeeds in producing a much more realistic road network.

\section{Machine-Assisted Map Editing}

To improve street map datasets, the inferred road network derived from iterative tracing and refinement steps must be incorporated into the existing road network map. Fully automated integration of the inferred road network is impractical: because the inferred roads may still include errors after refinement, adding all of the inferred roads to the map dataset would degrade the precision of the dataset.

Instead, we develop a human-in-the-loop map editing framework that enables human map editors to rapidly validate automatically inferred data~\cite{maid}. On initialization, our machine-assisted map editor builds an overlay containing the inferred road segments. Users interact with the overlay by left and right clicking to mark segments as correct or incorrect. Thus, rather than trace roads through a series of repeated clicks along the road, when a correct inferred segment already covers the road, users of the machine-assisted editor can rapidly add the road to the map with a single click on the inferred segment in the overlay.

Our map editor has two additional features to improve validation speed. First, the interface includes a prune button that, when pressed, executes a shortest-path based pruning algorithm to eliminate minor residential and service roads from the overlay, leaving only major arterial roads. This functionality is especially useful when mapping areas where the existing road network in the street map dataset has low coverage. In these areas, adding every missing road to the map may require substantial effort, but the quality of the map could already be improved significantly if major roads are incorporated. The pruning algorithm is effective at helping users focus on mapping these unmapped, major roads by reducing information overload.
%\sanjay{Maybe use the phrase "reduces information overload"}
%We identify major roads by determining whether a road segment frequently falls along the shortest path between distant points on the road network. Intuitively, because major roads offer fast connections between far apart locations, they should appear on shortest paths between such locations. Specifically, we first cluster the vertices of the road network to obtain cluster centers. Then, we compute shortest paths between cluster centers that are at least a minimum radius $R$ apart. For each shortest path, we trim a fixed distance from the ends of the path, which may correspond to minor roads near the source or destination of the path. We then mark every road in the remaining middle portion of the path as a major road. Then, the set of major roads is the union of the major roads marked on each shortest path computation.
We show an example pruning result in Figure \ref{fig:refinement}(b). Purple segments are pruned, leaving the yellow segments that correspond to major inferred roads.

Pruning is most useful in low-coverage areas. For high-coverage areas, we develop a teleport button that pans users to a connected component of inferred roads. In high-coverage areas, only a small number of roads are missing from the map, and identifying an unmapped road requires users to painstakingly scan the imagery one tile at a time. The teleport button eliminates this need, allowing users to jump to a group of missing roads and immediately begin mapping them.

\section{Future Work}

Although Mapster substantially reduces the workload of map creation and maintenance, the automatically inferred street maps still have more errors than maps created by professional map makers. Filling this gap requires advances in machine learning approaches for automatic map inference. Below, we detail several promising avenues to improve inference performance.

First, better neural network architectures can improve the performance of automatic mapping. So far, the design of the neural network architectures in Mapster are mostly inspired by the best practice in general computer vision tasks such as image segmentation and image classification. However, map inference tasks have unique characteristics such as the strong spatial correlation in the satellite imagery.
%\sanjay{Though iterative tracing was designed specifically for this task.}
Thus, improved neural network architectures that are specialized for map making tasks may yield higher accuracy. For example instead of using raw images or GPS traces as input, node and edge embeddings garnered from unsupervised tasks on extremely large map datasets could be used.

Second, end-to-end loss functions are a promising avenue to directly learn desired properties of the output road network. However, these desired properties, such as high precision and high recall over roads, generally can only be computed from the road network graph. As a result, the objectives are non-differentiable, since the graph can only be extracted from the probabilistic output of a machine learning model through non-differentiable functions. Thus, both prior work and our iterative tracing approach learn to build a graph indirectly: prior work minimizes the per-cell classification error, and in iterative tracing on satellite imagery we minimize the difference between the predicted angle of an untraced road and the correct angle on each tracing step. Nevertheless, end-to-end training has been shown to improve performance in other machine learning tasks, and could improve accuracy of automatic map creation and maintenance. Several techniques have been proposed for optimizing non-differentiable metrics. For example, reinforcement learning techniques can search for optimal policies under non-differentiable rewards. Alternatively, it may be possible to train an additional neural network which takes the intermediate map representation (e.g., road segmentation) as input and predicts the value of evaluation metrics. This additional neural network acts as a differentiable approximation of the evaluation metrics and can be leveraged to train a model to infer road networks that score highly on the metric.

In addition to potential improvements in the machine learning techniques, incorporating new data sources would also extend the capability of Mapster. In particular, two promising data sources are drone imagery and video from dashboard cameras. Drone imagery enables on-demand image sensing; for instance, if we find that the road structure in a region is unclear from satellite imagery and GPS data, we can reactively assign drones to collect aerial images over that region. Video from dashboard cameras enables inferring several additional street map features such as street names, business names, road signs, and road markers.

\section{Conclusion}

The world has approximately 64 million kilometers of roads and the road network is growing at a rapid pace as major countries like China, India, Indonesia, and Qatar gather economic momentum. Street maps are important. However, creating and maintaining street maps is very expensive and involves labor-intensive processes. As a result, although a lot of effort and money has been spent in maintaining street maps, today's street maps are still imperfect, and are frequently either incomplete or lag behind new construction. Mapster is a holistic approach for applying automation to reduce the work needed to maintain street maps. By incorporating automatic map inference with data refinement and a machine-assisted map editor, Mapster makes automation practical for map editing, and enables the curation of map datasets that are more complete and up-to-date at less cost.

\bibliographystyle{acm}
\bibliography{main}

\end{document}